\PassOptionsToPackage{hyperfootnotes=false}{hyperref}
\ifdefined\TMLRSubmission
  \documentclass[10pt]{article}
\else
  \documentclass{article}
\fi

\ifdefined\TMLRSubmission
  \usepackage{tmlr}
\else
  \usepackage{PRIMEarxiv}
  \usepackage[numbers,sort&compress]{natbib}
\fi
\usepackage[utf8]{inputenc}
\usepackage[T1]{fontenc}
\ifdefined\TMLRSubmission\else
  
\fi
\usepackage{amsmath,amssymb,amsfonts}
\usepackage{booktabs}
\usepackage{array}
\usepackage{tabularx}
\usepackage{longtable}
\usepackage{graphicx}
\usepackage{tikz}
\usetikzlibrary{arrows.meta,positioning,fit,backgrounds,shapes.geometric,calc}
\usepackage{microtype}
\usepackage{nicefrac}
\usepackage{url}
\usepackage[dvipsnames]{xcolor}

\definecolor{FigBlue}{HTML}{245B78}
\definecolor{FigTeal}{HTML}{188B87}
\definecolor{FigOrange}{HTML}{D88A16}
\definecolor{FigInk}{HTML}{263545}
\definecolor{FigMuted}{HTML}{647383}
\definecolor{FigViolet}{HTML}{805AA4}
\definecolor{FigRed}{HTML}{B74842}

\usepackage{flafter}
\usepackage{float}
\usepackage[section]{placeins}
\usepackage{hyperref}

\hypersetup{
  colorlinks=true,
  linkcolor=MidnightBlue,
  citecolor=MidnightBlue,
  urlcolor=MidnightBlue
}

\ifdefined\TMLRSubmission
\else
  \hypersetup{
    pdftitle={Replay-Gated Neural Execution: Decoupling Persistent Behavioral Specifications from Neural Realizations in Frozen Language Models},
    pdfauthor={Xianliang Zeng and Zhanzhan Zhao},
    pdfsubject={Replay-gated, state-indexed neural execution in frozen language models},
    pdfkeywords={activation steering, neural control, runtime assurance, transactional replay, frozen language models}
  }
\fi

\graphicspath{{media/}}
\newcolumntype{Y}{>{\raggedright\arraybackslash}X}

\newcommand{\Cref}{\mathcal{C}_{\mathrm{ref}}}
\newcommand{\Cfull}{\mathcal{C}_{\mathrm{full}}}
\newcommand{\Ccascade}{\mathcal{C}_{\mathrm{cascade}}}

\ifdefined\TMLRSubmission
\else
  \makeatletter
  \renewcommand{\@maketitle}{%
    \vbox{\hsize\textwidth\linewidth\hsize\centering
      {\LARGE\normalfont\@title\par}
      \vskip 14pt
      {\normalsize\normalfont\begin{tabular}[t]{c}\@author\end{tabular}\par}
      \vskip 18pt}}
  \makeatother
\fi

\title{
  \textbf{Replay-Gated Neural Execution:}\\[0.30em]
  {\Large Decoupling Persistent Behavioral Specifications from Neural Realizations
  in Frozen Language Models}
}

\ifdefined\TMLRSubmission
  \author{}
\else
  \author{
    Xianliang Zeng \qquad Zhanzhan Zhao \\
    The Chinese University of Hong Kong, Shenzhen \\
    Shenzhen, China \\
    \texttt{xianliangzeng@link.cuhk.edu.cn}
    \,\,\,
    \texttt{zhanzhanzhao@cuhk.edu.cn}
  }
\fi

\ifdefined\TMLRSubmission
  \newcommand{\ArtifactSentence}{An anonymized executable artifact and aggregate evidence package are included in the supplementary material.}
\else
  \newcommand{\ArtifactSentence}{A public reference implementation and aggregate frozen-evidence slice are available at \url{https://github.com/Messydrawing/replay-gated-neural-execution}.}
\fi

\begin{document}
\maketitle

\begin{abstract}
Input-conditioned neural interventions raise a runtime question: what persists when one behavioral specification admits multiple actions whose validity depends on execution state? We introduce replay-gated neural execution, separating five objects: a persistent behavioral predicate, its state-indexed certified realization set, a transient action witness, a budget-limited finder, and execution authorization. Candidates undergo isolated FP32/BF16 replay of the frozen model; commitment additionally requires a valid run audit.

Experiments on Qwen3-0.6B and SmolLM2-360M-Instruct establish distinct failure modes for these objects. Independent initializations yield distinct certified actions in all 24 tested fixed-state cells. Unchanged SmolLM2 witnesses remain certified in all 128 native states but only 66 of 384 off-diagonal transfers. All 767 archived Qwen witnesses replay successfully, yet a budget-limited finder misses one known-realizable cell in all three prespecified runs. Of 1,141 replay-submitted candidates, 174 fail item certification.

A frozen three-tier cascade uses these boundaries to reject uncertified proposals and escalate audit-valid search misses. On 256 previously sealed Qwen Fresh requests, 221 first certify at the lowest-cost tier and all 256 receive audited authorization, with no observed bypass. Relative to frozen full search, the median singleton search-and-certification cost ratio is 0.1055 and P95 is 1.3485, including failed tiers. Within the studied behavioral family on two small models, these results support state-indexed, set-valued execution semantics: specifications persist, search proposes witnesses, and replay certification plus run audit grants execution authority.
\end{abstract}

\ifdefined\TMLRSubmission
\else
\fi

\section{Introduction}
\label{sec:introduction}

Reusable neural interventions include function vectors, contrastive activation addition, and representation fine-tuning \citep{todd2024function,rimsky2024caa,wu2024reft}. More recent systems condition action construction, retrieval, strength, or placement on the current input or model state \citep{sun2025hypersteer,do2025dsem,li2026svf,hsu2026clas,weng2026finesteer,nelwan2026perinstance}. ATLAS uses a learned verifier over intermediate hidden states to decide, per example and reasoning step, whether and how strongly to apply steering \citep{nguyen2026atlas}; feedback-control and online-optimization methods adapt interventions during generation \citep{kong2024recontrol,nguyen2025pid,skifstad2026alqr}; and ObserverBench separates state estimation from the loss induced by a chosen action \citep{erramilli2026observerbench}. These lines study how an action is constructed, selected, or adjusted. An underexplored runtime question is what should persist, be searched for, and be authorized when one behavioral specification admits multiple neural realizations whose validity can depend on the current execution state.

Our central proposal is that a persistent behavioral specification is not a persistent neural action: $U\not\equiv a$. The persistent object remains $U$; under a frozen protocol $\Pi$ and paired execution state $\bar{s}$, its physical realization is represented by the state- and protocol-indexed certified realization set
\begin{equation}
  \mathcal{W}^{\Pi}_{U}(\bar{s})
  = \left\{a\in\mathcal{A}_{\mathrm{ABI}}:\operatorname{Cert}^{\mathrm{item}}_{\Pi}(U,\bar{s},a)=1\right\}.
  \label{eq:intro-certified-set}
\end{equation}
Here $\mathcal{A}_{\mathrm{ABI}}$ is the protocol's structural neural-action domain. The resulting semantics distinguishes the persistent specification $U$, its certified realization set $\mathcal{W}^{\Pi}_{U}(\bar{s})$, a transient witness $a$, a budget-limited finder $\gamma$, and execution authorization $\operatorname{Auth}_{\Pi}$. A finder proposes candidates; replay certification and a separate run audit determine whether one may alter model state.

The experiments expose why these objects cannot be collapsed. Repeated solving at fixed state produces distinct certified witnesses in all 16 tested Qwen cells and all 8 tested SmolLM2 cells. Direct cross-state replay retains certification in only 66 of 384 off-diagonal SmolLM2 transfers. Separately, all 767 archived Qwen reference witnesses remain replay-certified, while one known-realizable difficult cell is missed in all three runs of the frozen budget-limited finder. Thus certified identity differs from a particular witness, and protocol-relative certified realizability differs from operational rediscovery.
For a finder $\gamma$ with finder-step budget $B$ and run identity or randomness $\xi$, this distinction is expressed as
\begin{equation}
  \mathcal{W}^{\Pi}_{U}(\bar{s}) \neq \varnothing
  \quad\not\Rightarrow\quad
  \operatorname{Rediscover}_{\Pi}(\gamma,B,U,\bar{s};\xi)=1.
  \label{eq:existence-not-rediscovery}
\end{equation}

This separation has an operational consequence. Across 1,141 candidates submitted to replay, 174 fail the item certificate, so candidate generation alone cannot confer execution authority. The resulting runtime uses a frozen cascade $\gamma_1\rightarrow\gamma_2\rightarrow\gamma_3$: a tier that obtains no certified candidate within its budget triggers escalation only if its run audit is valid, and every tier remains subject to the same ReplayCert and run audit. On previously sealed Qwen Fresh contexts, 221 of 256 requests certify at the lowest-cost tier and all 256 are ultimately authorized. Relative to frozen full search, the median singleton search-and-certification cost ratio is $0.1055$, with P95 $1.3485$. We introduce this state-indexed, set-valued, replay-gated execution semantics and show that certified realizability, witness generation, and execution authority have empirically distinct failure modes.

\paragraph{Scope of the present claims.}
The present study is restricted to runtime realization and authorization for a supplied behavioral predicate. Utility construction, post-freeze installation, capability composition, model-class universality, bare-generation latency, and formal safety for arbitrary future trajectories are outside the scope of the present claims.

\begin{figure}[ht!]
  \centering
  \resizebox{\linewidth}{!}{
\begin{tikzpicture}[
  font=\footnotesize\sffamily,text=FigInk,
  flow/.style={-{Latex[length=1.9mm]},draw=FigInk!75,line width=.75pt},
  box/.style={rounded corners=3pt,align=center,line width=.65pt,inner sep=5pt},
  sub/.style={font=\scriptsize\sffamily,text=FigMuted,align=center}]
  \fill[FigViolet!5,rounded corners=4pt] (0,.15) rectangle (3.05,5.93);
  \fill[FigBlue!4,rounded corners=4pt] (3.40,.15) rectangle (7.05,5.93);
  \fill[FigTeal!4,rounded corners=4pt] (7.40,.15) rectangle (12.00,5.93);
  \fill[FigInk!3,rounded corners=4pt] (12.35,.15) rectangle (16.45,5.93);
  \foreach \x/\label in {.20/Specify,3.60/Propose,7.60/Certify,12.55/Authorize}{
    \node[anchor=west,font=\footnotesize\sffamily\bfseries,text=FigMuted] at (\x,5.63) {\label};
  }
  \node[box,draw=FigViolet!60,fill=white,minimum width=2.70cm,minimum height=1.05cm] (u) at (1.52,4.30)
    {\textbf{Semantic target $U$}\\[3pt]{\scriptsize desired behavior}};
  \node[box,draw=FigInk!30,fill=white,minimum width=2.70cm,minimum height=1.05cm] (state) at (1.52,2.15)
    {\textbf{Snapshot $\bar s$}\\[3pt]{\scriptsize hidden state / KV / RNG}};
  \node[box,draw=FigBlue!60,fill=white,minimum width=3.20cm,minimum height=1.05cm] (finder) at (5.22,4.30)
    {\textbf{Budgeted search $\gamma_k$}\\[3pt]
    {\scriptsize\textcolor{FigBlue}{Tier 1}\ $\to$\ \textcolor{FigTeal}{Tier 2}\ $\to$\ \textcolor{FigOrange}{Tier 3}}};
  \node[box,draw=FigBlue!60,fill=white,minimum width=3.20cm,minimum height=1.05cm] (a) at (5.22,2.15)
    {\textbf{Candidate action $a$}\\[3pt]{\scriptsize residual-stream writes}};
  \draw[flow] (u.east) -- (finder.west);
  \draw[flow] (finder.south) -- (a.north);
  \draw[flow,draw=FigMuted] (state.north) -- (1.52,3.18) -- node[above,sub]{redacted telemetry $o$} (4.08,3.18) -- (4.08,0 |- finder.south);
  \node[box,draw=FigTeal!55,fill=white,minimum width=4.20cm,minimum height=2.35cm] (replay) at (9.70,2.15) {};
  \node[font=\footnotesize\sffamily\bfseries] at (9.70,3.04) {Frozen-model replay};
  \foreach \y/\layer in {2.58/7,2.20/14,1.82/21}{
    \fill[FigInk!7,rounded corners=1pt] (7.98,\y-.13) rectangle (9.66,\y+.13);
    \node[anchor=west,font=\tiny\sffamily] at (8.02,\y) {L\layer};
    \foreach \x in {8.82,9.13,9.44}{\fill[FigBlue] (\x,\y) circle[radius=1.55pt];}
  }
  \node[box,draw=FigTeal!45,fill=FigTeal!5,minimum width=1.36cm,minimum height=.43cm,font=\scriptsize\sffamily] at (10.77,2.48) {FP32};
  \node[box,draw=FigTeal!45,fill=FigTeal!5,minimum width=1.36cm,minimum height=.43cm,font=\scriptsize\sffamily] at (10.77,1.90) {BF16};
  \node[sub] at (9.70,1.29) {9 write slots; frozen weights};
  \draw[flow] (a.east) -- (replay.west);
  \draw[flow,draw=FigMuted] (state.south) -- (1.52,.60) -- node[below,sub]{restore before replay} (9.70,.60) -- (replay.south);
  \node[box,draw=FigTeal!65,fill=white,minimum width=4.20cm,minimum height=1.05cm] (cert) at (9.70,4.30)
    {\textbf{ReplayCert}\\[3pt]{\scriptsize valid for this target and state}};
  \draw[flow,draw=FigTeal] (replay.north) -- (cert.south);
  \node[box,draw=FigInk!35,fill=white,minimum width=3.45cm,minimum height=1.05cm] (audit) at (14.40,4.30)
    {\textbf{Run audit}\\[3pt]{\scriptsize trajectory integrity}};
  \draw[flow,draw=FigMuted] (finder.north) -- (5.22,5.14) -- node[below,sub,inner sep=1pt]{search trajectory $T$} (14.40,5.14) -- (audit.north);
  \node[box,draw=FigTeal!65,fill=white,minimum width=3.45cm,minimum height=1.05cm] (auth) at (14.40,2.15)
    {\textbf{Execution authorization}\\[3pt]{\scriptsize certificate + audit must pass}};
  \draw[flow,draw=FigTeal] (cert.east) -- (12.17,4.30) |- (auth.west);
  \draw[flow] (audit.south) -- (auth.north);
  \node[box,draw=FigBlue,fill=FigBlue,text=white,minimum width=3.45cm,minimum height=.68cm] (commit) at (14.40,.60)
    {\textbf{Commit action $a$}};
  \draw[flow] (auth.south) -- (commit.north);
\end{tikzpicture}}
  \caption{Authority-separated neural execution. A semantic target $U$ guides budgeted search for a candidate $a$. ReplayCert tests that candidate by isolated FP32/BF16 replay of the frozen model from the original state. Execution requires both item certification and a passing run audit. The Qwen schematic shows nine declared residual-write slots. Audit-valid misses escalate or abstain; integrity failures stop (Figure~\ref{fig:proposal-funnel}).}
  \label{fig:overview}
\end{figure}
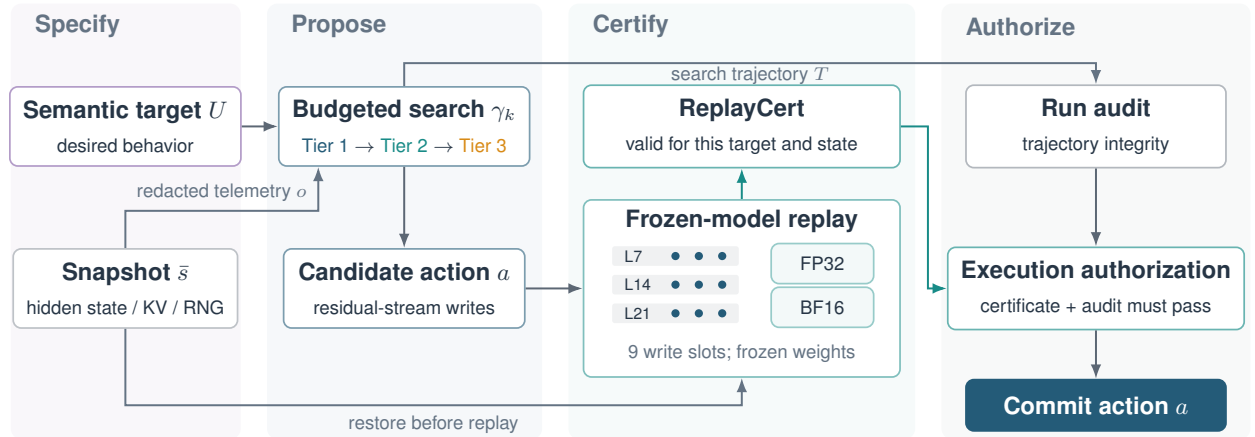

\section{Related Work}
\label{sec:related-work}

\subsection{From Fixed Interventions to Context-Conditioned Neural Control}

Fixed activation interventions show that intermediate state can be an actionable inference-time interface: Activation Engineering, CAA, Function Vectors, ITI, and ReFT construct or learn reusable representation changes while leaving the backbone frozen \citep{turner2023activation,rimsky2024caa,todd2024function,li2023iti,wu2024reft}. Representation Engineering, affine surgery, and activation scaling broaden this interface beyond one hand-crafted direction \citep{zou2023representation,singh2024surgery,stoehr2024scaling}. These temporary actions differ from persistent weight editing \citep{meng2022rome,meng2023memit}; distributed representations and superposition also caution against equating one semantic object with one neuron or direction \citep{park2023linear,elhage2022superposition,huben2024sae,templeton2024scaling}.

More recent systems make direction, magnitude, layer, or composition input dependent, spanning state-conditioned synthesis, retrieval, local gradient fields, adaptive strength, routed bases, and conditional proposal distributions \citep{sun2025hypersteer,do2025dsem,li2026svf,hsu2026clas,weng2026finesteer,vogels2026ids,nelwan2026perinstance,han2026steer2adapt,zeng2026more,wang2026diffusteer}. ATLAS uses a learned verifier over intermediate hidden states to decide, per example and reasoning step, whether and how strongly to apply steering \citep{nguyen2026atlas}. In the present runtime, learned, verifier-guided, and optimized actions remain proposals. Membership in $\mathcal{W}^{\Pi}_{U}(\bar{s})$ is assigned only by transactionally isolated replay of the frozen model under the current state and protocol.

Non-identifiability and attribution results motivate this separation: geometrically different steering vectors can be behaviorally indistinguishable, apparent effects can follow answer encodings, and efficacy need not imply attribution specificity \citep{venkatesh2026nonidentifiability,gao2026crossencoding,luo2026steercheck}. Causal conclusions from activation patching likewise depend on perturbation, site, and metric \citep{zhang2023patching}. The fixed-state multi-witness experiment operationalizes a narrower form of this ambiguity under one frozen certification contract.

\subsection{Online Synthesis, Feedback Control, and Runtime Admission}

Online neural control provides a second precedent: feedback methods adapt interventions instead of applying one fixed open-loop vector \citep{kong2024recontrol,cheng2024linearcontrol,nguyen2025pid,skifstad2026alqr}. ObserverBench fixes the intervention problem and reports estimation accuracy separately from the loss of the selected action \citep{erramilli2026observerbench}. Selective prediction formalizes rejection \citep{elyaniv2010selective,geifman2019selectivenet,lee2024selectivegeneration}. MERA calibrates whether and how strongly to intervene and may abstain, whereas CAP provides context-adaptive risk and abstention policies for language and vision-language models \citep{hedstrom2025mera,tayebati2025cap}. Neural Simplex, shielding, and BRT-Align provide broader precedents for separating a performance or proposal mechanism from runtime assurance or safety intervention \citep{phan2020neuralSimplex,alshiekh2018shielding,karnik2025brtalign}. ReplayCert addresses empirical item admission by isolated FP32/BF16 execution of the actual frozen model, followed by run authorization for the current restored state.

\subsection{Research Gap and Positioning}

Prior work has made neural interventions increasingly conditional on inputs and internal states, and has separately introduced verifier-guided selection, feedback control, abstention, and runtime assurance. These advances primarily address how an intervention is proposed, selected, adjusted, or guarded. They do not by themselves resolve what should constitute the persistent object when one behavioral specification admits multiple certified neural realizations.

We study that distinction directly. Rather than identify a persistent behavioral specification $U$ with one neural action $a_U$, we represent its protocol-relative realization by the state-indexed set $\mathcal{W}^{\Pi}_{U}(\bar{s})$, and separately model a transient witness $a$, an incomplete finder $\gamma$, and execution authorization. The experiments test whether these distinctions correspond to empirically different failure modes: fixed-state non-uniqueness, cross-state membership change, certified existence without operational rediscovery, and proposal without certification or authorization. To our knowledge, adjacent methods focus principally on proposal, selection, control, or assurance rather than making this persistent-object distinction itself the empirical question. Extended method-by-method positioning appears in Appendix~\ref{app:related}.

\section{Problem Formulation: Operational Set-Valued Neural Execution Semantics}
\label{sec:formulation}

\subsection{Protocol-Relative Precision-Paired Execution State and Redacted Observation}
Let $M$ denote a frozen language model. Each backbone is paired with its own frozen execution protocol $\Pi_M$, structural action domain $\mathcal{A}_{\mathrm{ABI},M}$, and precision-paired state $\bar{s}_M$. The behavioral predicate $U$ is model-independent at the specification level; the model index applies to the execution state, structural action domain, effect map, and certified realization set $\mathcal{W}^{\Pi_M}_{U,M}(\bar{s}_M)$. All definitions below are instantiated separately for each backbone, and we suppress $M$ when no ambiguity arises. An execution episode is represented by
\begin{equation}
  \bar{s}=\left(s^{32},s^{\mathrm{bf}}\right).
\end{equation}
Both components bind the same model checkpoint, input, tokenization, generation boundary, random-number identity, and protocol version, but are restored from separate FP32 and BF16 snapshots. Thus $\bar{s}$ is a certification/evaluation pair, not one numerical state represented twice; its hidden tensors are distinct numerical objects. The protocol fixes which $s^{\mathrm{serve}}\in\{s^{32},s^{\mathrm{bf}}\}$ is exposed to the finder and permits only redacted telemetry $o=\phi(s^{\mathrm{serve}})$ under the frozen redaction map $\phi$; the withheld fields are specified in Appendix~\ref{app:protocol}. The structural domain $\mathcal{A}_{\mathrm{ABI}}$ fixes layers, write positions, and tensor shapes but does not itself impose numerical certification gates. An action $a\in\mathcal{A}_{\mathrm{ABI}}$ is replayed and certified on the paired state. The primary instantiation uses frozen Qwen3-0.6B \citep{yang2025qwen3} and fixed residual-stream ports at Transformer layers \citep{vaswani2017attention}; attention, MLP, and all other model weights remain unchanged.

\subsection{Persistent Behavioral Specification and the State-Indexed Effect Map}
Let $\mathcal{Y}^{32}$ and $\mathcal{Y}^{\mathrm{bf}}$ denote the spaces containing the scored behavioral and effect quantities used by $U$ under FP32 and BF16 replay. The internal term ``Utility'' denotes a typed predicate of behavioral success, not a scalar reward to maximize:
\begin{equation}
  U:\mathcal{Y}^{32}\times\mathcal{Y}^{\mathrm{bf}}\longrightarrow\{0,1\},
  \qquad
  \mathcal{S}_{U}=U^{-1}(\{1\}).
\end{equation}
Neither $U$ nor its acceptance set $\mathcal{S}_{U}$ changes with execution state. For $p\in\{32,\mathrm{bf}\}$, the component map $F^{p}_{s^{p}}:\mathcal{A}_{\mathrm{ABI}}\to\mathcal{Y}^{p}$ sends a structurally admissible action to the effect/output measured by replay from restored state $s^{p}$. State indexing is carried by the resulting dual-precision action-to-effect map
\begin{equation}
  \mathbf{F}^{\Pi}_{\bar{s}}(a)
  =
  \left(F^{32}_{s^{32}}(a),F^{\mathrm{bf}}_{s^{\mathrm{bf}}}(a)\right)
  \in\mathcal{Y}^{32}\times\mathcal{Y}^{\mathrm{bf}}.
\end{equation}

\subsection{Certified Realization Set and Transient Witness}
Structural validity is represented by $a\in\mathcal{A}_{\mathrm{ABI}}$. Separately, let $K^{\mathrm{item}}_{\Pi}(U,\bar{s},a)\in\{0,1\}$ denote the frozen protocol's numerical execution contract, including action-energy and per-write limits, finiteness, and protocol-applicable invariance or exact-zero checks. Top-1, target-effect peak, and robust margin belong to the behavioral acceptance set $\mathcal{S}_{U}$ rather than being redefined inside $K^{\mathrm{item}}_{\Pi}$. Section~\ref{sec:runtime} and Appendix~\ref{app:protocol} specify the concrete gates. Single-action certification is
\begin{equation}
  \operatorname{Cert}^{\mathrm{item}}_{\Pi}(U,\bar{s},a)
  =
  \mathbf{1}\!\left[\mathbf{F}^{\Pi}_{\bar{s}}(a)\in\mathcal{S}_{U}\right]
  \land K^{\mathrm{item}}_{\Pi}(U,\bar{s},a).
\end{equation}
Accordingly,
\begin{equation}
  \mathcal{W}^{\Pi}_{U}(\bar{s})
  =
  \left\{a\in\mathcal{A}_{\mathrm{ABI}}:
  \operatorname{Cert}^{\mathrm{item}}_{\Pi}(U,\bar{s},a)=1\right\}.
  \label{eq:certified-set}
\end{equation}
The persistent behavioral specification is $U$; $\mathcal{W}^{\Pi}_{U}(\bar{s})$ is its protocol-relative certified realization set under the paired execution state. Any $a\in\mathcal{W}^{\Pi}_{U}(\bar{s})$ is termed a \emph{certified witness} of $U$ at $\bar{s}$ under protocol $\Pi$. The witness is transient even when $U$ persists. We make no assumptions of manifold geometry, smoothness, connectivity, or global identifiability for this set.

\subsection{Incomplete Finder and Operational Rediscovery}
Because the search process may generate candidates adaptively from permitted intermediate feedback, its primary object is the ordered trajectory
\begin{equation}
  T_{\gamma,B}(U,o;\xi)
  =
  \bigl((a_1,z_1),\ldots,(a_T,z_T)\bigr),
  \qquad T\leq B,
\end{equation}
where $\xi$ denotes run randomness or run identity and $z_t$ is search feedback available to the finder under the protocol. Here $B$ is the maximum number of protocol-defined finder steps, not the economic cost metric reported in Section~\ref{sec:design}. The candidate set is only the projection $C(T)=\{a_1,\ldots,a_T\}$. Operational rediscovery is defined as
\begin{equation}
  \operatorname{Rediscover}_{\Pi}(\gamma,B,U,\bar{s};\xi)
  =
  \mathbf{1}\!\left[
  \exists t\leq T:\,
  a_t\in\mathcal{W}^{\Pi}_{U}(\bar{s})
  \right].
  \label{eq:rediscovery-success}
\end{equation}
The state $\bar{s}$ enters $\operatorname{Rediscover}$ only through the evaluator-side membership test; it is not an additional observation available to the finder. Because we do not assume that $\gamma$ is complete under budget $B$, certified existence does not guarantee operational rediscovery:
\begin{equation*}
  \mathcal{W}^{\Pi}_{U}(\bar{s})\neq\varnothing
  \nRightarrow
  \operatorname{Rediscover}_{\Pi}(\gamma,B,U,\bar{s};\xi)=1.
\end{equation*}

\subsection{Run Audit and Execution Authorization}
Single-action validity is not equivalent to validity of the run process. Let $a^{\star}\in\mathcal{A}_{\mathrm{ABI}}\cup\{\bot\}$ denote the action selected for commitment, where $a^{\star}=\bot$ means that the run selected no certified candidate. The run audit
\begin{equation}
  \operatorname{Audit}^{\mathrm{run}}_{\Pi}(T,\bar{s},a^{\star})
  \in\{0,1\}
\end{equation}
checks restoration from the original snapshot, isolation of hidden-state, key--value (KV) cache, and random-number-generator (RNG) state between candidates, prohibited information access, and the snapshot, Utility, and protocol bindings of the trajectory. When $a^{\star}\neq\bot$, it additionally checks the selected-action hash, final replay, and commitment binding; these action-specific checks are inapplicable when $a^{\star}=\bot$. For a selected action $a\in\mathcal{A}_{\mathrm{ABI}}$, execution authorization is
\begin{equation}
  \operatorname{Auth}_{\Pi}(U,\bar{s},a,T)
  =
  \operatorname{Cert}^{\mathrm{item}}_{\Pi}(U,\bar{s},a)
  \land
  \operatorname{Audit}^{\mathrm{run}}_{\Pi}(T,\bar{s},a).
  \label{eq:execution-auth}
\end{equation}
The executor commits only actions for which $\operatorname{Auth}_{\Pi}=1$. If a finite search returns no certified candidate and $\operatorname{Audit}^{\mathrm{run}}_{\Pi}(T,\bar{s},\bot)=1$, the runtime may escalate to the next frozen tier; exhaustion of Tier3 yields explicit abstention. A run-audit or integrity failure instead terminates fail closed without commitment and is not recoverable by switching finders.

These definitions complete the separation between behavioral specification, certified realization, witness discovery, and execution authority. Section~\ref{sec:runtime} instantiates this factorization as a transactional replay runtime, and Section~\ref{sec:design} specifies the experimental cohorts and adjudication rules used to test the corresponding failure modes.

\section{Replay-Gated Neural Execution: Physical Runtime}
\label{sec:runtime}

\subsection{Backbone-Specific Physical ABIs}
For each backbone, search, replay, and commitment are restricted to one backbone-specific frozen physical ABI. Here, ``physical ABI'' refers only to the corresponding frozen experimental intervention contract; it is not a completed post-freeze Neural Capability ABI. For Qwen3-0.6B, an action performs nine residual-stream writes at Transformer layers 7, 14, and 21, with three fixed write positions at each layer and one $1{,}024$-dimensional residual vector per write. A single action can therefore be represented as
\begin{equation}
  a=(a_{\ell,w})_{\ell\in\{7,14,21\},\,w\in\{1,2,3\}},
  \qquad a_{\ell,w}\in\mathbb{R}^{1{,}024},
\end{equation}
with a physical coordinate dimension of $3\times3\times1{,}024=9{,}216$. All model weights, attention parameters, and MLP parameters remain frozen; actions alter the current episode's intermediate state only through the declared ports. Fixed ports, positions, and shapes define the structural domain $\mathcal{A}_{\mathrm{ABI}}$, while finiteness and the frozen energy and write-norm limits in Appendix~\ref{app:protocol} are checked by $K^{\mathrm{item}}_{\Pi}$. This separates structural admissibility from numerical execution constraints.

The $9{,}216$ dimensions are the physical coordinates of the final Qwen action; individual finders need not perform the same unconstrained search over all coordinates. Tier1, Tier2, and Tier3 use different frozen candidate-generation dynamics, but all produce actions under the Qwen ABI and are adjudicated by the same ReplayCert. SmolLM2 uses the separate $8{,}640$-dimensional ABI in Section~\ref{sec:smollm-design}. Each $\mathcal{W}^{\Pi_M}_{U,M}(\bar{s})$ is therefore relative to one model and protocol.

\subsection{Transactional Singleton Replay}
Every candidate starts from the same original snapshot:
\begin{equation*}
  \text{snapshot}\rightarrow\text{candidate replay}\rightarrow
  \text{discard}\rightarrow\text{restore}\rightarrow\text{next candidate}.
\end{equation*}
Candidate trials exist only to obtain certification evidence; their temporary execution states do not become the final commitment state. After selection, the system restores the original snapshot again and executes the same action, binding the final commitment to the certified action identity rather than to mutable state produced during search. Modified hidden state, KV/cache state, and random-number-generator state never flow between candidate trials.

\subsection{ReplayCert and Execution Authorization}
The action-adjudication component of ReplayCert implements $\operatorname{Cert}^{\mathrm{item}}_{\Pi}$. An action belongs to $\mathcal{W}^{\Pi}_{U}(\bar{s})$ only when separately restored replays at both precisions satisfy Top-1, target-effect peak, and robust margin; all single-action energy and write contracts; finiteness of action and output; and the exact-zero reference path where applicable, together with equivariance under the frozen candidate-list transformation family. ``Certified by replay'' denotes empirical admission under this frozen contract, not formal verification.

The run-adjudication component implements $\operatorname{Audit}^{\mathrm{run}}_{\Pi}$. It verifies that each candidate trial restores the original snapshot, that no hidden-state, KV-cache, or RNG-state leakage occurs between candidates, that the finder reads no protocol-prohibited information, and that the snapshot, Utility, and protocol identities are bound consistently to the trajectory. If an action is selected, its hash and final replay are additionally bound to the commitment; an audit-valid search miss instead uses $a^{\star}=\bot$ and has no action-specific commitment check. A hard violation, information-path violation, or evidence-binding mismatch fails the run audit. Passing the single-action certificate is therefore still insufficient for commitment; the executor commits only when $\operatorname{Auth}_{\Pi}=1$ in Eq.~\ref{eq:execution-auth}.

\subsection{Certified Adaptive Dynamics Cascade}
\begin{table}[!htbp]
  \caption{Frozen cascade structure. Exact internal phase lengths for Tier3 are listed in the appendix.}
  \label{tab:cascade}
  \centering
  \footnotesize
  \renewcommand{\arraystretch}{1.12}
  \begin{tabularx}{\linewidth}{l Y >{\raggedright\arraybackslash}p{.115\linewidth} Y Y}
    \toprule
    Tier & Candidate generation & Update budget & Visible information & Escalation or termination \\
    \midrule
    Tier1 & Protocol-visible $K=8$ pullback warm start followed by a short full-space certified solve & $\leq32$ & $U,o$, and protocol-permitted $z_t$ & Escalate after an audit-valid search miss; fail closed on an invalid run \\
    Tier2 & Augmented curvature and relinearization & $\leq48$ & Same frozen interface & Escalate after an audit-valid search miss; fail closed on an invalid run \\
    Tier3 & Singleton WitnessSearch matching the full baseline & Frozen full lifecycle & Same frozen interface & Abstain after audit-valid search exhaustion; fail closed on an invalid run \\
    \bottomrule
  \end{tabularx}
\end{table}
Each tier re-instantiates its frozen initializer. Optimizer state, hidden/KV/RNG state, search trajectory, and candidate actions from the preceding tier are not passed forward. The re-instantiated Tier3 lifecycle matches that of the full operational baseline. Only a candidate with $\operatorname{Auth}_{\Pi}=1$ is committed; a search miss escalates, an invalid run fails closed, and exhaustion of Tier3 returns abstention.

The protocol-visible $K=8$ pullback supplies only Tier1's initial action; the subsequent short solve updates the complete physical-action schedule before ReplayCert admission. Useful search geometry need not itself be a sufficient realization space.

The frozen budgets, lifecycles, and key implementation settings for each tier appear in the appendix. The paper relies only on three protocol-level properties: tiers increase in cost, each tier is reinitialized independently, and all final candidates are adjudicated by the same ReplayCert and execution-authorization rule.

\section{Experimental Design and Frozen Evidence Protocol}
\label{sec:design}

\subsection{Qwen3 Cohorts and Isolation}
The basic unit of analysis is a pairing between one frozen execution context and one Utility, hereafter called a cell. Each complete context contains 64 predeclared Utilities, so four contexts form $4\times64=256$ test cells. Every cell uses the same-sized frozen candidate list and the same behavioral, precision, and physical execution contracts. Experiments differ only in the context cohort, the candidate-finding process, and whether previously established reference witnesses may be read.

\paragraph{Context construction and split.}
The contexts are synthetic prompt templates from a preconstructed design pool rather than samples selected after observing cascade results. An archived context record is $\tau=(\texttt{context\_id},\texttt{family},\texttt{prompt},\texttt{address\_prefix})$: the address prefix is a literal prompt prefix that fixes the token-generation boundary, and, under the frozen execution protocol, the model and tokenizer produce the paired FP32/BF16 execution snapshots at that boundary. Thus, contexts vary in prompt form and the resulting hidden state; candidate registry, tokenizer, model, ABI, and the 64 Utilities remain fixed. The source pool contains 64 contexts in eight format families, with eight contexts per family. The split unit is the entire context, so all 64 Utilities associated with one context remain in the same cohort. Before execution, a fixed seed and SHA-256 ordering assign four families to CascadePilot and four to Frozen Fresh, then select one context per family; the freeze receipt verifies both context- and family-disjointness. Reference Stability and retired contexts are excluded by committed context identities. The commitments record context identifiers, families, selection ranks, seed, and source-pool hash; raw prompt payloads are withheld from the distributed artifact.

Specifically, the Z64 contract defines 64 predeclared two-digit continuation strings (\texttt{00}--\texttt{63}) as $q_0,\ldots,q_{63}$, with tokenization and generation boundaries frozen. Scores use the mean-centered, length-normalized sequence log probability $s_j^p(a)$ defined in Appendix~\ref{app:protocol}. Given a context $\tau$ and target index $j$, Utility $U_j$ requires $q_j$ to be Top-1 among all 64 items under both FP32 and BF16 replay, with a target-versus-runner-up score margin of at least $0.2$ at each precision. In addition, relative to the zero-action baseline, the score increase of $q_j$ must be maximal among the 64 items. An experimental cell is therefore $(\tau,U_j)$, and ``$256/256$'' means that all 256 such objectives across four frozen contexts receive an action that passes both the dual-precision behavioral gate and the physical contract. This synthetic typed task adjudicates the neural execution interface; it is not a benchmark of natural-language capability.

We use four experimental cohorts with distinct scientific roles (Table~\ref{tab:populations}). Reachability Public establishes the reachability of low-energy dual-precision physical realizations under the current ABI. Reference Stability contains $12\times64=768$ fixed cells: 767 have certified reference witnesses and form a known-realizable reference set, while one is an explicit abstention fixed before execution. This cohort separates ``a known certified realization remains valid'' from ``the current finder can rediscover a realization.'' The identities of 23 difficult cells were frozen before formal reruns to construct a difficult rediscovery test, not to estimate the overall failure rate of the finder. CascadePilot provides a one-time qualification and cost gate for the cascade. After it passes, the cascade policy, tier budgets, ReplayCert, effect coordinate system, and cohort identity are no longer modified. Only then is the previously sealed Frozen Fresh cohort first read and used as the primary new-context evaluation of the frozen policy.

The Unseen Utility cohort remains sealed throughout these experiments and contributes neither results nor model, finder, certificate, or threshold selection. The study begins after a Utility is supplied and does not test post-freeze late binding.

\begin{table}[H]
  \caption{Qwen3 experimental cohorts and their roles.}
  \label{tab:populations}
  \centering
  \small
  \begin{tabularx}{\linewidth}{l Y Y}
    \toprule
    Cohort & Size & Purpose \\
    \midrule
    Reachability Public & $4\times64$ & Physical reachability and cross-precision gating \\
    Reference Stability & $12\times64=768$ (767 reference + 1 abstention) & Witness replay and difficult-cell rediscovery \\
    CascadePilot & $4\times64$ & One-time qualification and cost gate \\
    Frozen Fresh & $4\times64$ & New-context evaluation of the frozen policy \\
    Unseen Utility & Sealed & Outside the scope of this paper \\
    \bottomrule
  \end{tabularx}
\end{table}
\FloatBarrier

\subsection{SmolLM2 Replication Cohorts}
\label{sec:smollm-design}
The second backbone is frozen SmolLM2-360M-Instruct, with 32 layers and hidden width 960. Its ABI writes at layers 8, 16, and 24 during three protocol-fixed phases, giving $3\times3\times960=8{,}640$ coordinates. The model, tokenizer, ports, scorer, projection, and the same physical energy limits are frozen; the exact revision and implementation settings appear in Appendix~\ref{app:protocol}.

Three SmolLM2 panels have distinct roles. The fixed-state restart diagnostic uses two retired DesignPublic contexts and eight independent initializations per cell. The Fresh execution panel uses four context- and family-disjoint contexts with 32 frozen Utilities each and compares the frozen K8 cascade, a zero-initialized short solve, and an optimized early-stop full solve. Cross-state transfer then freezes those 128 native actions and transactionally replays each one in all four target contexts without search, adaptation, or action modification. No panel reads a teacher, parent action, or checkpoint. Appendix~\ref{app:protocol} gives the exact model revision, restart settings, Fresh Utility selection rule, search budgets, timing boundary, arm order, and cross-state source-action rule; frozen execution identities are recorded in the accompanying evidence archive. The executable real-model miniature uses separate retired contexts and is not part of these scientific populations.

\subsection{Three Coverage Objects}
We separately report frozen-reference coverage, full-operational coverage, and cascade coverage:
\begin{equation}
  \Cref,\qquad \Cfull,\qquad \Ccascade.
\end{equation}
$\Cref$ is the set of cells shown to be ``known realizable'' by frozen reference witnesses; it is not a complete oracle for every realizable cell in the action space. $\Cfull$ is the set certified by the frozen full operational solver, which is likewise an operational baseline rather than a realizability oracle. $\Ccascade$ is the set certified by the frozen cascade under the same ReplayCert and authorization rules. Cascade coverage relative to the reference set is
\begin{equation}
  R_{\mathrm{ref}}^{\mathrm{cascade}}
  = \frac{|\Ccascade\cap\Cref|}{|\Cref|}.
\end{equation}

\subsection{Behavioral, Admission, and Cost Metrics}
We report dual-precision behavioral gates, execution-authorization audits, contract-violating commitments, abstentions, first-certifying tier, normalized search-and-certification cost, and paired-replay-equivalent cost. For the Qwen normalized-cost analysis, the active-Utility window $m$ is the number of Utilities within one frozen context that share one telemetry/local-response acquisition. For each context and each of 64 cyclic starting indices, the window contains $m$ consecutive Utility indices modulo 64. Shared acquisition cost is counted once, while the search-and-certification cost of each Utility remains included individually. Thus, $m$ is not a batch size of independent contexts and does not change the certification rule for any cell. The primary windows are $m=1$ and $m=4$; $m=2,8,64$ are secondary analyses. The SmolLM2 panel instead reports measured wall-clock and CUDA time, replay count, optimizer evaluations, and peak allocated memory.

The exact metric, quantile, bootstrap, and cost-accounting rules are frozen in Appendix Table~\ref{tab:metrics}. Top-1, effect peak, and robust margin are evaluated separately in FP32 and BF16; cohort P99 energy is a population admission gate rather than an item-certificate condition. The primary economic windows are $m=1,4$. Paired request-level cascade/full ratios include failed tiers, state restoration, and dual-precision replay; model loading and evidence hashing are excluded, so the metric is not bare-model end-to-end latency. Full formulas and resampling caveats appear in Appendix~\ref{app:protocol}.

\subsection{Cohort-Level Scientific Adjudication}
Let $\mathcal{D}$ denote a fixed cohort of context--Utility cells together with their replay and cost records. The deterministic adjudicator
\begin{equation}
  \operatorname{Accept}^{\mathrm{cohort}}_{\Pi}(\mathcal{D})\in\{0,1\}
\end{equation}
checks population-level P99 energy, coverage, tier distribution, cost statistics, protocol-defined bootstrap bounds, and evidence completeness under $\Pi$. These cohort quantities adjudicate a scientific population but do not participate in single-action membership in $\mathcal{W}^{\Pi}_{U}(\bar{s})$. A fixed cell lacking a certified reference witness remains in the cohort denominator and replay audit.

\subsection{Evidence Validity}
Scientific \textsc{Pass}, scientific \textsc{Fail}, engineering \textsc{Invalid}, and a \textsc{Sealed} cohort are distinct terminal states. Hash, environment, cohort, and closeout checks are reported in the appendix as credibility controls rather than novelty contributions.

\subsection{Reproducibility and Artifact Boundary}
\ArtifactSentence{} The release separates four objects: archived aggregate Qwen evidence; an executable real-model SmolLM2 miniature on retired contexts; source and hash manifests for the reference replay, authorization, and cascade abstractions; and raw formal Qwen context/witness payloads, which are not distributed. The miniature exercises the same object boundaries but cannot regenerate the archived Qwen study. The repository documents evidence identities, verification commands, and reproduction limits; Appendix~\ref{app:audit} summarizes the audit chain.

\section{Experimental Results}
\label{sec:results}

Results follow the four empirical claims stated in the introduction. Original protocol labels and the complete claim--evidence ledger remain in Appendix Table~\ref{tab:evidence-ledger}.

\subsection{Certified Realizations Exist and Are Non-Unique at Fixed State}
The cross-precision solver produces certified actions for $256/256$ public Qwen cells with zero hard violations. Maximum energy is approximately $0.01738$, below the $0.021$ single-action limit, while cohort P99 is approximately $0.01675$, below the separate $0.018$ admission threshold. Widespread physical unreachability of the studied ports is therefore not supported on this cohort.

Non-uniqueness is tested without changing $(U,\bar{s})$. On Qwen, all 128 actions from eight independent initializations over 16 cells pass the full certificate, and all 448 within-cell pairs are numerically distinct (relative-$\ell_2$ range $0.0225340$--$1.1948683$; minimum cosine similarity $0.2861448$). SmolLM2 similarly certifies all $64/64$ runs over eight fixed cells, with all $224/224$ pairs distinct (relative-$\ell_2$ range $0.206739$--$0.734279$). These panels reject one archived action as the unique canonical target; they do not estimate population frequency or global realization-set geometry. Appendix Figure~\ref{fig:reference-stability} records the full diagnostic.

\subsection{Certified Membership Changes Across Execution States}
We replay each of 128 certified SmolLM2 source-context actions in every target context without search or adaptation. Native replay remains $128/128$, whereas only $66/384=17.19\%$ of off-diagonal transfers remain certified. All 12 ordered context pairs contain a failure, and $121/128$ source cells fail in at least one non-native target. Physical validity survives all 384 transfers and target-effect peak survives 351, but only 123 retain Top-1, 69 retain robust margin, and 66 retain the full certificate. Failure therefore arises mainly in context-sensitive candidate competition rather than action legality. Nonzero overlap remains, so the result establishes material state indexing, not disjoint sets or unique state attribution.

\begin{figure}[!htbp]
  \centering\sffamily
  \begin{minipage}[t]{0.43\linewidth}
    \centering
    \textbf{\footnotesize (a) Certified source--target transfers}\par\medskip
    \begin{tikzpicture}[x=1.07cm,y=0.84cm,font=\footnotesize\sffamily]
      \node at (2,5.02) {Target context};
      \foreach \x/\lab in {0/XML,1/DB Row,2/Log,3/Chat}{
        \node[font=\scriptsize\sffamily] at (\x+0.5,4.40) {\lab};
      }
      \node[rotate=90] at (-1.85,2) {Source context};
      \foreach \y/\lab in {3/XML,2/DB Row,1/Log,0/Chat}{
        \node[anchor=east] at (-0.17,\y+0.5) {\lab};
      }
      \foreach \x/\y/\cnt in {
        0/3/32,1/3/5,2/3/7,3/3/2,
        0/2/6,1/2/32,2/2/3,3/2/2,
        0/1/11,1/1/5,2/1/32,3/1/2,
        0/0/8,1/0/8,2/0/7,3/0/32}{
        \pgfmathsetmacro{\shade}{100*\cnt/32}
        \fill[FigBlue!\shade] (\x,\y) rectangle (\x+1,\y+1);
        \ifnum\cnt=32
          \node[white,font=\footnotesize\sffamily\bfseries] at (\x+.5,\y+.5) {\cnt};
        \else
          \node at (\x+.5,\y+.5) {\cnt};
        \fi
      }
      \node at (2,-.45) {Full certificates per 32 transfers};
    \end{tikzpicture}
  \end{minipage}\hfill
  \begin{minipage}[t]{0.55\linewidth}
    \centering
    \textbf{\footnotesize (b) Off-diagonal gate pass counts}\par\medskip
    \begin{tikzpicture}[x=0.0104cm,y=0.74cm,font=\footnotesize\sffamily]
      \node at (192,5.70) {384 transfers; each gate evaluated separately};
      \foreach \lab/\cnt/\y in {Physical contract/384/4.6,Effect peak/351/3.6,Top-1/123/2.6,{Margin $\geq0.2$}/69/1.6,Full certificate/66/0.6}{
        \fill[FigBlue!75] (0,\y-0.27) rectangle (\cnt,\y+0.27);
        \node[anchor=east] at (-9,\y) {\lab};
        \node[anchor=west] at (\cnt+7,\y) {\cnt};
      }
      \draw[black!55,line width=.4pt] (0,0) -- (384,0);
      \foreach \tick in {0,100,200,300,384}{
        \draw[black!55,line width=.4pt] (\tick,0) -- (\tick,-.10);
        \node[anchor=north] at (\tick,-.13) {\tick};
      }
      \node at (192,-1.02) {Off-diagonal transfers (count)};
    \end{tikzpicture}
  \end{minipage}
  \normalfont
  \caption{SmolLM2 cross-state transfer of unchanged witnesses. (a) Rows are source contexts and columns are target contexts, with 32 Utilities per source. XML: XML notice; DB Row: database row; Log: log excerpt; Chat: chat transcript. All $128/128$ native replays certify, versus $66/384$ off-diagonal transfers. (b) Each gate is evaluated on all 384 off-diagonal transfers; bars are not cumulative filtering stages. Physical legality persists, while Top-1 and robust-margin qualification change across contexts.}
  \label{fig:smollm-transfer}
\end{figure}
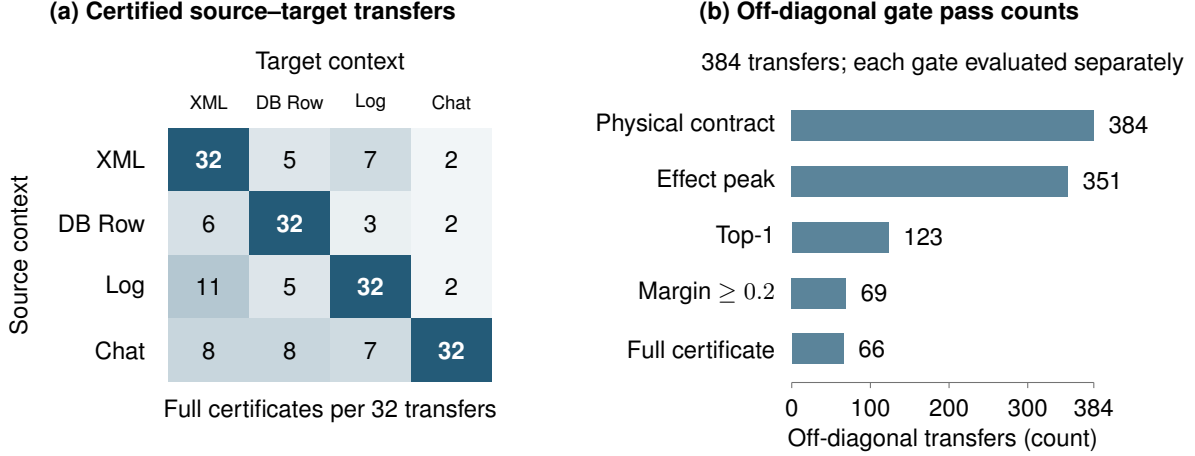

\subsection{Search, Certification, and Authorization Are Distinct}
Reference Stability Stage A transactionally replays 767 singleton-native reference witnesses without search; all $767/767$ pass, while the cell lacking a reference remains an explicit abstention. Stage B runs three budget-limited singleton rediscovery lifecycles for each of 23 difficult cells selected before execution. Twenty-two cells certify in all three repeats, but $(\text{context }9,U37)$ abstains in all three without hard or nonfinite failure. Its Stage A witness still certifies, establishing $\mathcal{W}^{\Pi}_{U}(\bar{s})\neq\varnothing$ while the three specified $\operatorname{Rediscover}_{\Pi}$ evaluations are zero. This is a counterexample to treating one finite finder as a realizability oracle, not a population failure-rate estimate or randomized basin study.

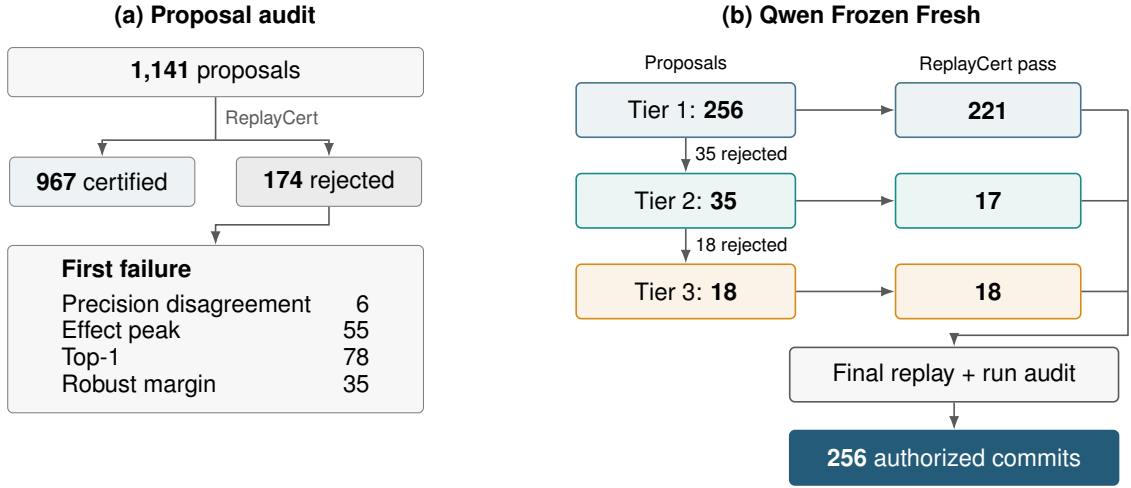
\begin{figure}[!htbp]
  \centering\sffamily
  \begin{minipage}[t]{0.40\linewidth}
    \centering
    \textbf{\footnotesize (a) Proposal audit}\par\medskip
    \begin{tikzpicture}[font=\footnotesize\sffamily,
      box/.style={draw=black!50,rounded corners=2pt,align=center,minimum height=.65cm,inner sep=5pt},
      arr/.style={-{Latex[length=1.8mm]},draw=black!65,line width=.6pt}]
      \node[box,fill=black!3,minimum width=5.5cm] (all) at (2.8,0)
        {\textbf{1,141} proposals};
      \node[box,fill=FigBlue!8,minimum width=2.45cm] (eligible) at (1.3,-1.45)
        {\textbf{967} certified};
      \node[box,fill=black!8,minimum width=2.45cm] (reject) at (4.3,-1.45)
        {\textbf{174} rejected};
      \coordinate (split) at (2.8,-.90);
      \draw[black!65,line width=.6pt] (all.south) -- node[right,font=\scriptsize\sffamily]{ReplayCert} (split);
      \draw[arr] (split) -| (eligible.north);
      \draw[arr] (split) -| (reject.north);
      \node[box,fill=black!3,minimum width=5.5cm] (reasons) at (2.8,-3.40) {
        \begin{tabular}{@{}lr@{}}
          \multicolumn{2}{@{}l@{}}{\textbf{First failure}}\\[3pt]
          Precision disagreement & 6\\
          Effect peak & 55\\
          Top-1 & 78\\
          Robust margin & 35
        \end{tabular}};
      \draw[arr] (reject.south) -- (4.3,-2.02) -| (reasons.north);
    \end{tikzpicture}
  \end{minipage}\hfill
  \begin{minipage}[t]{0.58\linewidth}
    \centering
    \textbf{\footnotesize (b) Qwen Frozen Fresh}\par\medskip
    \begin{tikzpicture}[font=\footnotesize\sffamily,
      box/.style={rounded corners=2pt,align=center,minimum height=.72cm,inner sep=5pt,line width=.6pt},
      proposal/.style={box,minimum width=2.9cm},
      certified/.style={box,minimum width=2.45cm},
      arr/.style={-{Latex[length=1.8mm]},draw=black!65,line width=.6pt}]
      \node[font=\scriptsize\sffamily] at (1.5,.60) {Proposals};
      \node[font=\scriptsize\sffamily] at (5.5,.60) {ReplayCert pass};
      \node[proposal,draw=FigBlue!85,fill=FigBlue!8] (t1) at (1.5,0)
        {Tier 1: \textbf{256}};
      \node[proposal,draw=FigTeal,fill=FigTeal!9] (t2) at (1.5,-1.20)
        {Tier 2: \textbf{35}};
      \node[proposal,draw=FigOrange,fill=FigOrange!10] (t3) at (1.5,-2.40)
        {Tier 3: \textbf{18}};
      \node[certified,draw=FigBlue!85,fill=FigBlue!8] (c1) at (5.5,0)
        {\textbf{221}};
      \node[certified,draw=FigTeal,fill=FigTeal!9] (c2) at (5.5,-1.20)
        {\textbf{17}};
      \node[certified,draw=FigOrange,fill=FigOrange!10] (c3) at (5.5,-2.40)
        {\textbf{18}};
      \foreach \i in {1,2,3}{
        \draw[arr] (t\i.east) -- (c\i.west);
      }
      \draw[arr] (t1.south) -- node[right,font=\scriptsize\sffamily]{35 rejected} (t2.north);
      \draw[arr] (t2.south) -- node[right,font=\scriptsize\sffamily]{18 rejected} (t3.north);
      \coordinate (join1) at (7.35,0);
      \coordinate (join2) at (7.35,-1.20);
      \coordinate (join3) at (7.35,-2.40);
      \foreach \i in {1,2,3}{\draw[black!65,line width=.6pt] (c\i.east) -- (join\i);}
      \draw[black!65,line width=.6pt] (join1) -- (join3);
      \node[box,draw=black!65,fill=black!3,minimum width=4.35cm] (audit) at (5.05,-3.50)
        {Final replay + run audit};
      \draw[arr] (join3) -- (7.35,-2.98) -| (audit.north);
      \node[box,draw=FigBlue,fill=FigBlue,text=white,minimum width=4.35cm] (auth) at (5.05,-4.60)
        {\textbf{256} authorized commits};
      \draw[arr] (audit) -- (auth);
    \end{tikzpicture}
  \end{minipage}
  \normalfont
  \caption{Proposal, certification, and authorization. (a) Audited Qwen and SmolLM2 events, with mutually exclusive rejection counts (Appendix~\ref{app:audit}). (b) Audit-valid rejections escalate; selected certified actions require final replay and run audit before commitment. All 256 requests commit, with zero observed authorization bypasses. Node sizes are schematic.}
  \label{fig:proposal-funnel}
\end{figure}

We also audit every tier-boundary or diagnostic-arm proposal submitted to transactional replay; optimizer states never sent to ReplayCert are excluded. Of 1,141 proposal events, 967 are item-certificate eligible and 174 are rejected under the mutually exclusive first-failure rule in Appendix~\ref{app:audit}. No rejection arises from a physical-contract, information-path, candidate-permutation, or nonfinite-value failure. In the Qwen Frozen Fresh cascade, ReplayCert rejects 53 intermediate candidates and escalates their cells before the selected final actions pass the run audit and yield $256/256$ authorized commitments with zero observed bypass (Figure~\ref{fig:proposal-funnel}).

\subsection{Replay-Gated Execution Preserves Coverage at Lower Cost}
\paragraph{Reference and Pilot.}
On the frozen 767-cell reference set, the cascade certifies $767/767$ whereas the frozen full operational solver certifies $766/767$; first certification occurs at Tier1/Tier2/Tier3 for $709/33/25$ cells. On CascadePilot, both methods certify $256/256$ and every cascade certification occurs at Tier1. The Pilot is process-integrity evidence that the low-cost tier suffices on that cohort; it does not establish adaptive escalation.

\paragraph{Frozen Fresh.}
After freezing policy, budgets, certificate, run audit, effect coordinates, and cohort identity, both methods certify $256/256$ previously sealed cells. First-certifying-tier counts are $221/17/18$: $35/256=13.67\%$ of requests escalate beyond Tier1 and $18/256=7.03\%$ first certify only at Tier3. Context-level counts are $36/12/16$, $58/5/1$, $63/0/1$, and $64/0/0$. Committed-energy P99 and maximum are both approximately $0.017749$, satisfying the distinct $0.018$ cohort and $0.021$ single-action gates; exact values appear in Appendix~\ref{app:tables}.

\begin{figure}[htbp]
  \centering
  \includegraphics[width=\linewidth]{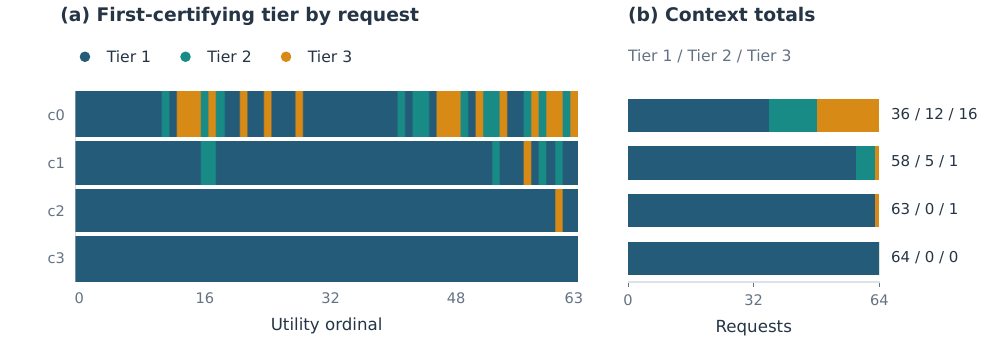}
  \caption{Frozen Fresh routing at matched granularity. Rows align the 64 Utility requests in each context with their Tier 1/2/3 totals. Context 0 accounts for 28 of the 35 requests that escalate; all 256 requests ultimately certify. Historical cohort aggregates remain in Table~\ref{tab:evidence-ledger}.}
  \label{fig:tier-distribution}
\end{figure}

\paragraph{Cost and second-backbone replication.}
At singleton window $m=1$, the paired request-level cascade/full cost ratio has median $0.1055$ and P95 $1.3485$. Typical requests are substantially cheaper, but the cascade does not uniformly dominate full search: its difficult tail can be more expensive. Table~\ref{tab:cost} includes failed tiers under the common accounting boundary. Appendix Figures~\ref{fig:fresh-diagnostics} and~\ref{fig:cost-frontier} show how request-level costs vary by first-certifying tier and how shared acquisition is amortized across active-Utility windows.

\begin{table}[htbp]
  \caption{Frozen Fresh normalized search-and-certification cost. LCB95 is the protocol-defined 95\% bootstrap lower bound over frozen cyclic windows; it is not an i.i.d. population interval.}
  \label{tab:cost}
  \centering
  \small
  \begin{tabular}{rrrrr}
    \toprule
    Active $U$ ($m$) & $1-\operatorname{med}(r_i)$ & LCB95 & $\operatorname{P95}(r_i)$ & \shortstack{Paired-replay-equivalent\\cost reduction} \\
    \midrule
    1  & $89.45\%$ & $88.98\%$ & 1.3485 & $85.59\%$ \\
    2  & $91.37\%$ & $90.90\%$ & 1.0973 & $88.13\%$ \\
    4  & $92.91\%$ & $92.28\%$ & 0.9434 & $90.54\%$ \\
    8  & $93.32\%$ & $92.42\%$ & 0.9037 & $91.44\%$ \\
    64 & $88.23\%$ & $84.24\%$ & 0.6479 & $86.44\%$ \\
    \bottomrule
  \end{tabular}
\end{table}

On four group-disjoint SmolLM2 Fresh contexts, the cascade, zero-initialized short solve, and optimized early-stop full solve each certify $128/128$ cells. Cascade wall and CUDA time are $0.3923\times$ optimized full search (339.5 versus 865.5 seconds), with 840 versus 6,400 transactional replays and 4.666 versus 7.813 GB peak allocated memory. Its K8 warm start is 5.495\% faster than the matched zero-initialized short solve, though it uses more replays and memory. Every cell first certifies at Tier1, so this panel replicates low-cost certified execution, not adaptive routing. Cost comparisons are against frozen or optimized operational baselines, not bare-model inference or external steering systems.

\section{Discussion}
\label{sec:mechanistic}

\subsection{From an Action Object to a State-Indexed Execution Problem}
For the behavioral family studied here, the persistent object is the predicate $U$ that specifies success. Certification eligibility of a concrete action is indexed by execution state and protocol; the behavioral specification does not own one persistent vector. An action in $\mathcal{W}^{\Pi}_{U}(\bar{s}_1)\cap\mathcal{W}^{\Pi}_{U}(\bar{s}_2)$ is valid in both paired states, so state indexing does not imply a unique state identity. The SmolLM2 transfer matrix establishes material state indexing with partial overlap: most off-diagonal transfers fail, while $66/384$ remain certified.

\subsection{Proposal Probability Is Not Certified Membership}
When a learned or stochastic finder induces a proposal distribution, its probability mass and the certified realization set answer different questions. High proposal probability for an action $a$ does not imply $a\in\mathcal{W}^{\Pi}_{U}(\bar{s})$; membership is determined only by the frozen replay certificate at the paired execution state. Conversely, a certified witness can receive negligible or zero probability under a particular proposer and therefore remain operationally undiscovered. Thus, proposal support, protocol-relative membership, and execution authorization must not be identified with one another. The deterministic analogue is the same: reachability by one search trajectory is neither necessary nor sufficient for certified membership.

Replay certification is also distinct from attribution. It records that a concrete witness satisfies a declared behavioral and physical contract in one restored execution state; it does not establish a semantically unique, encoding-invariant, or causally privileged internal representation.

\subsection{Adaptive Cost Allocation without a Realizability Oracle}
Replay separates candidate generation from authorization: uncertified candidates are never committed, audit-valid misses escalate or abstain, and integrity failures terminate fail closed. The frozen reference, full operational solver, and cascade sets are reported separately because a deterministic miss does not prove that the certified realization set is empty.

The cascade allocates computation across incomplete proposal mechanisms. A miss at a cheap tier triggers the frozen escalation path because every tier has proposal authority only; later tiers remain subject to the same admission and authorization rule. Frozen Fresh retains $256/256$ coverage while concentrating additional effort on the $35/256$ requests that escalate beyond Tier1. Frozen escalation and a common gate therefore preserve system-level certified coverage despite incomplete individual tiers.

\section{Limitations and Falsifiability}
\label{sec:limitations}

\begin{itemize}
  \item \textbf{Generality, Utility, and interface scope.} Evidence comes from two small frozen backbones and backbone-specific residual-write ABIs. SmolLM2 replicates fixed-state non-uniqueness, low-cost execution, and state dependence, but not adaptive routing. Utilities are synthetic Z64 predicates; natural capability descriptions, Unseen Utility installation, composition, arbitrary identifier remappings, and encoding-invariant attribution remain untested \citep{gao2026crossencoding,luo2026steercheck}.

  \item \textbf{Comparison and deployment cost scope.} External methods have not been compared under a matched action domain, energy limit, budget, and replay contract. Reported reductions use frozen full WitnessSearch as the comparator and exclude loading and hashing; bare-model latency remains unmeasured, and singleton P95 exceeds full search.

  \item \textbf{Finite geometry and empirical assurance.} Restart panels establish numerical non-uniqueness but not prevalence or topology; four-context transfer establishes overlap and state dependence but not model-wide intersection frequency; repeated rediscovery from one analytic start measures lifecycle stability, not basin coverage. Authorization records and zero observed bypass are empirical evidence, not a formal safety proof. The released artifact omits raw formal Qwen payloads and the historical target runner.
\end{itemize}

\section{Broader Impact}
\label{sec:impact}

Separating proposals from execution authority may support auditable, fail-closed inference-time control, but can also make behavioral manipulation more reliable. Releases should therefore avoid high-risk specifications, alignment-bypass objectives, and unattended automatic commitment.

\section{Conclusion}
\label{sec:conclusion}

Under the studied protocols, certified realizations are non-unique at fixed state and are produced by incomplete search. The evidence supports treating the persistent predicate $U$, its state- and protocol-indexed certified realization set $\mathcal{W}^{\Pi}_{U}(\bar{s})$, a transient witness $a$, the budget-limited finder $\gamma$, and final authorization $\operatorname{Auth}_{\Pi}$ as distinct runtime objects. Neither a proposal nor finder success confers execution authority. Unseen Utility, post-freeze installation, removal, and composition remain separate questions for future neural-capability ABIs.

\begingroup
\small
\ifdefined\TMLRSubmission
  \bibliographystyle{tmlr}
\else
  \bibliographystyle{unsrt}
\fi
\bibliography{references}
\endgroup

\appendix
\ifdefined\TMLRSubmission
  \clearpage
\else
  \par\vspace{1.5\baselineskip}
\fi

\section{Full Protocol and Frozen Gates}
\label{app:protocol}
The primary Qwen runtime uses Qwen3-0.6B with frozen model and tokenizer identities. An action contains nine $1{,}024$-dimensional residual-stream writes at layers 7, 14, and 21. The hard bridge-energy, answer-energy, per-write, and total-energy limits in the single-action certificate are $0.006$, $0.015$, $0.05$, and $0.021$, respectively. The P99 threshold of $0.018$ is checked separately by the cohort admission function and does not determine membership in $\mathcal{W}^{\Pi}_{U}(\bar{s})$. Tier1 has a budget of at most 32 updates, and Tier2 at most 48. Tier3 uses a frozen analytic $\rightarrow97\rightarrow388\rightarrow$ relinearization lifecycle identical to that of the full operational comparison baseline. Every tier re-instantiates a fresh initializer and inherits neither the preceding tier's search trajectory or candidates nor any mutated state.

The SmolLM2 instantiation uses the frozen \texttt{SmolLM2-360M-Instruct} backbone from \texttt{HuggingFaceTB}. The bound revision is \texttt{a10cc1512eabd3dde888204e902eca88bddb4951}; writes occur at layers 8, 16, and 24, with hidden width 960 and three fixed write phases. Its fixed-state restart panel fixes Utilities $\{0,21,42,60\}$ and seeds $\{101,211,307,419,523,631,743,857\}$. Each restart uses a fresh candidate-blind legal initialization and Adam with learning rate $0.004$ for at most 64 updates. The physical and cohort limits match those above.

For the SmolLM2 Fresh execution panel, the frozen Utility subset is
\begin{equation*}
\begin{split}
\mathcal U_{\mathrm{Fresh}}=\{&0,4,5,6,8,9,11,12,14,15,20,21,22,23,25,28,31,32,\\
&36,37,38,40,42,43,45,47,49,51,54,55,58,60\}.
\end{split}
\end{equation*}
These are the first 32 ordinals under ascending SHA-256 order of the strings \texttt{30903002:ordinal}, then numerically sorted; the context and family split is independently frozen with seed $30903001$. Tier1 uses the protocol-visible $K=8$ pullback warm start followed by at most 32 full-space Adam updates, while zero-short begins at the zero action with the same 32-update cap. The optimized-full arm uses the frozen analytic initializer and, only when required, an early-stopping fallback capped at 97 updates. A single worker evaluates the three arms sequentially; their order is rotated across the four contexts as $(K8,Z,F)$, $(Z,F,K8)$, $(F,K8,Z)$, and $(K8,F,Z)$. Before each timed arm, one unmeasured zero-action replay is performed for each precision and scorer. Wall time spans the arm call under \texttt{time.perf\_counter}; CUDA time uses events with synchronization immediately before and after the arm, and peak allocated memory is reset at arm start.

For cross-state transfer, the source action for a source-context--Utility cell is the exact float32 action stored for that cell by the certified frozen-K8-cascade arm in the Fresh panel. Its byte-level SHA-256 identity is recorded, and the unchanged action is transactionally replayed from each target context's original snapshot in both precisions and with both scorers. No search, adaptation, or action modification is permitted during transfer.

\begin{table}[ht]
  \caption{Frozen metric and statistical definitions.}
  \label{tab:metrics}
  \centering
  \small
  \begin{tabularx}{\linewidth}{l Y}
    \toprule
    Metric & Frozen definition \\
    \midrule
    Top-1 & Target ranks first among all 64 candidates, evaluated separately in FP32 and BF16 \\
    Effect peak & Target-score effect is maximal relative to the zero action \\
    Robust margin & Target exceeds the runner-up by at least $0.2$ at each precision \\
    Exact zero & Empty Utility or abstention keeps the hard write path exactly zero \\
    Permutation & Equivariance error under the frozen candidate-list transformation family is at most $10^{-6}$; this does not refer to search-generation order \\
    Cohort P99 & Nearest-rank/upward quantile over committed actions; threshold $\leq0.018$, separate from item certification \\
    Economic window & Primary $m\in\{1,4\}$; secondary $m\in\{2,8,64\}$ \\
    Bootstrap & $10{,}000$ frozen-window resamples with seed $924117$; protocol-defined 95\% lower bound for the median reduction \\
    Cost accounting & Paired-replay-equivalent units calibrated on RTX 3090, including failed tiers, restored dual-precision replay, escalation, shared acquisition, and per-Utility segments \\
    \bottomrule
  \end{tabularx}
\end{table}

For cell $i$, both methods use the same frozen accounting boundary:
\begin{align}
  C_i &= C_i^{\mathrm{search}}+C_i^{\mathrm{failed\ tiers}}+C_i^{\mathrm{restore}}
  +C_i^{\mathrm{FP32}}+C_i^{\mathrm{BF16}}+C_i^{\mathrm{shared}}+C_i^{\mathrm{segment}},\\
  r_i &= \frac{C_i^{\mathrm{cascade}}}{C_i^{\mathrm{full}}}.
  \label{eq:cost-ratio}
\end{align}
The reported P95 is the upward empirical quantile of $\{r_i\}$, not a ratio of two cost quantiles, and the median reduction is $1-\operatorname{median}_i(r_i)$. For $m=1,4$, the frozen economic gates require at least a 25\% median reduction in normalized search-and-certification cost, at least a 30\% median reduction in paired-replay-equivalent cost, a nonnegative protocol-defined 95\% bootstrap lower bound, and $\operatorname{P95}_i(r_i)\leq1.35$. Each context/Utility/tier is physically run once; independently calibrated shared-acquisition and per-Utility segments are assembled deterministically into 64 cyclic windows. Costs of failed tiers remain included. Because those windows overlap within a context, the bootstrap is a frozen resampling summary rather than an independent-and-identically-distributed population-level confidence interval.

For precision $p\in\{32,\mathrm{bf}\}$, let $q_{j,1:L}$ be the frozen tokenization of candidate $q_j$, with $L=3$ under the Z64 registry. The score used for ranking, margins, and zero-action-relative effects is the mean-centered, length-normalized sequence log probability
\begin{equation*}
  \widetilde{s}^{p}_j(a)
  =\frac{1}{L}\sum_{t=1}^{L}\log p^{p}_{M}
  \!\left(q_{j,t}\mid \tau,q_{j,<t};a\right),
  \qquad
  s^{p}_j(a)
  =\widetilde{s}^{p}_j(a)-\frac{1}{64}\sum_{k=0}^{63}\widetilde{s}^{p}_k(a).
\end{equation*}
It is therefore not a raw final-token logit. FP32 and BF16 scores are computed separately from their restored snapshots.

The action component of ReplayCert implements $\operatorname{Cert}^{\mathrm{item}}_{\Pi}$. It performs FP32 and BF16 replay from mutually isolated restored snapshots and requires each precision to satisfy target Top-1, target-effect peak, and a robust margin of at least $0.2$. It also checks all single-action limits, finite outputs, exact-zero behavior, and equivariance under the frozen candidate-list transformation family at tolerance $10^{-6}$. The run component implements $\operatorname{Audit}^{\mathrm{run}}_{\Pi}$. Candidates are replayed one at a time from the same original hidden-state/KV-cache/RNG snapshot, all mutated state is discarded between trials, and prohibited information access, hard violations, and snapshot/Utility/protocol bindings are verified. If a certified action is selected, its action hash and final replay are additionally bound to the commitment; when no action is selected, $a^{\star}=\bot$ and only trajectory-level integrity checks apply. Before commitment, the selected action receives one final replay from the original snapshot. Execution is authorized only when both the single-action certificate and run audit pass.

Cohort identity and access discipline are frozen as follows. Reachability is public ($4\times64$). Reference Stability is archived evidence retired from further development use ($12\times64=768$, comprising 767 certified reference witnesses and one explicit abstention). CascadePilot ($4\times64$) is group-disjoint from the Reference Stability cohort. Frozen Fresh is opened only after the policy, budgets, certificate, effect coordinate system, and cohort identity are frozen ($4\times64$). Unseen Utility remains sealed. Statistical recombination follows the rules in Table~\ref{tab:metrics}, including bootstrap seed 924117.

Terminal states are not interchangeable. \textsc{Pass} and \textsc{Scientific-Fail} are valid scientific outcomes. \textsc{Invalid} denotes an identity, hash, environment, nonfinite-value, unexpected output on the standard-error stream (\texttt{stderr}), or integrity failure and cannot support a scientific conclusion. \textsc{Sealed} denotes an unopened cohort. A fixed cell lacking a certified reference witness remains in the fixed cohort denominator and replay audit.

\ifdefined\TMLRSubmission\clearpage\fi
\section{Complete Experimental Table}
\label{app:tables}
\begingroup
  \footnotesize
  \renewcommand{\arraystretch}{1.12}
  \begin{longtable}{@{}>{\raggedright\arraybackslash}p{.22\linewidth} >{\raggedright\arraybackslash}p{.17\linewidth} >{\raggedright\arraybackslash}p{\dimexpr.40\linewidth-6\tabcolsep\relax} >{\raggedright\arraybackslash}p{.21\linewidth}@{}}
    \caption{Frozen evidence ledger and the role of each block.}\label{tab:evidence-ledger}\\
    \toprule
    Evidence block & Cohort & Frozen result & Role in the claim \\
    \midrule
    \endfirsthead
    \multicolumn{4}{l}{\tablename~\thetable{} (continued)}\\
    \toprule
    Evidence block & Cohort & Frozen result & Role in the claim \\
    \midrule
    \endhead
    \midrule
    \multicolumn{4}{r}{\emph{Continued on next page}}\\
    \endfoot
    \bottomrule
    \endlastfoot
    Protocol H1: reachability & $4\times64$ & $256/256$; P99 $\approx.01675$; max $\approx.01738$ & Port reachability \\
    Protocol H2: non-uniqueness & 16 cells $\times$ 8 initializations & 16/16 each have 8 distinct certified solutions; 448 pairs; relative $\ell_2$ min $0.0225340$, max $1.1948683$ & Actions are non-unique at fixed cell \\
    Protocol H3: Reference Stage A & 767 cells & $767/767$; minimum margins $.20184/.20464$ & Frozen-witness validity \\
    Protocol H3: Reference Stage B & 23 preselected difficult cells $\times3$ & 22 cells $3/3$; one cell $0/3$ & Existence counterexample; no population-rate estimate \\
    Retrospective cascade & 767 cells & $767/767$ versus full solver $766/767$; $709/33/25$ & Full solver is not a realizability criterion \\
    Protocol H4: CascadePilot & $4\times64$ & $256/256$; $256/0/0$; no authorization bypass & Tier1 sufficient on Pilot \\
    Protocol H4: Frozen Fresh & $4\times64$ & $256/256$; $221/17/18$; no authorization bypass & Frozen adaptive execution \\
    SmolLM2 fixed-state restart panel & 8 cells $\times$ 8 initializations & $64/64$ certified; $224/224$ pairs distinct & Second-backbone non-uniqueness replication \\
    SmolLM2 Fresh execution & $4\times32$ & All three arms $128/128$; cascade/full time ratio $0.3923$; 840 vs. 6,400 replays & Second-backbone low-cost execution; no adaptive routing observed \\
    SmolLM2 cross-state transfer & 128 native + 384 off-diagonal & $128/128$ native; $66/384$ off-diagonal certified & State-indexed membership with overlap \\
    Proposal/certificate event audit & 1,141 replay-submitted events & 967 item-certified; 174 rejected & Proposal and item membership differ operationally \\
    Executable real-model miniature & 2 contexts $\times$ 4 Utilities & $8/8$ native; $4/8$ off-diagonal descriptive & Executable artifact, not Qwen formal reproduction \\
  \end{longtable}
\endgroup

For Reference Stability, the only fixed cell without a certified reference witness is $(\text{context }9,U39)$, which remains an abstention. The only operational-rediscovery cell with $0/3$ is $(\text{context }9,U37)$. The context-level Tier1/Tier2/Tier3 rows for Frozen Fresh are $36/12/16$, $58/5/1$, $63/0/1$, and $64/0/0$, with escalation rates of $43.75\%$, $9.38\%$, $1.56\%$, and $0\%$. Committed-energy P99 and maximum are $0.017749001533$ and $0.017749001754$. The minimum FP32 margins by context are $0.2004$, $0.2012$, $0.2007$, and $0.2001$; the corresponding minimum BF16 margins are $0.2006$, $0.2003$, $0.2041$, and $0.2067$. The complete active-window cost table appears in Table~\ref{tab:cost}. Attempts in failed tiers remain included in cost rather than being removed by conditional analysis.

\section{Audit and Reproducibility Chain}
\label{app:audit}
Protocol, code, model, tokenizer, environment, cohort, result, and closeout artifacts bind identity through SHA-256 manifests. Pilot and Fresh adjudication checks cohort identity, expected row counts, finite numerical fields, zero information-path violations, and terminal closeout status. Frozen Fresh closes with \texttt{FROZEN\_FRESH\_PASS}, \texttt{VALID\_CLOSEOUT}, and \texttt{CLOSEOUT\_AUDIT\_PASS}. These controls establish provenance and guard against post hoc changes to cohorts or thresholds. They are credibility controls, not additional algorithmic contributions. Exact hashes and machine-specific environment identifiers are recorded in the accompanying evidence archive; the public repository exposes the released model, source, and environment identities.

For the proposal audit, each replayed candidate receives exactly one mutually exclusive label using the frozen first-failure precedence: nonfinite or integrity failure; physical-contract or information-path failure; FP32/BF16 qualification disagreement; target-effect-peak failure; Top-1 failure; robust-margin failure; otherwise item-certificate eligible. This ordering prevents a candidate that fails several gates from contributing to several rejection categories. Authorization remains a later decision because it additionally requires a valid trajectory audit and selection under the frozen policy.

\ifdefined\TMLRSubmission\clearpage\fi
\section{Additional Realization and Search Evidence}
\label{app:geometry}
\begin{figure}[ht]
  \centering
  \includegraphics[width=\linewidth]{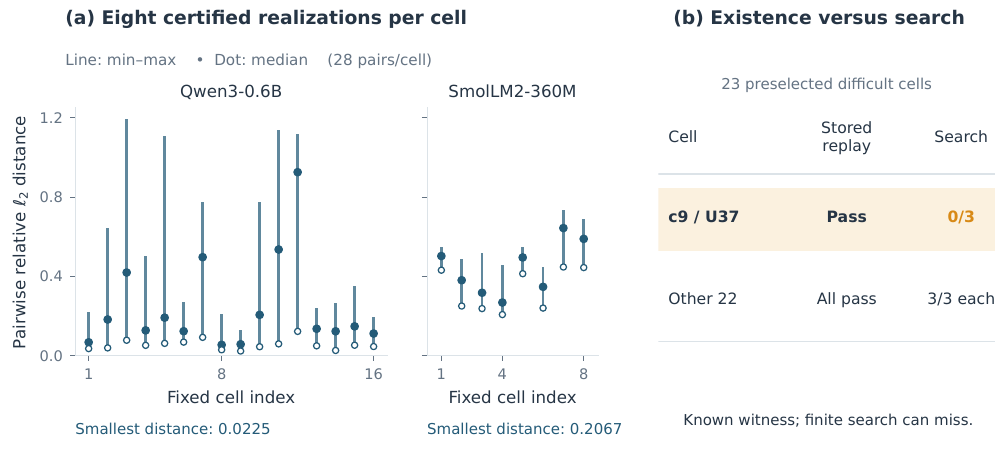}
  \caption{Certified non-uniqueness and finite-search failure. (a) For each fixed cell, a line spans the minimum to maximum of 28 pairwise distances among eight certified restarts; a filled dot marks the median and an open dot the minimum. These are descriptive ranges, not confidence intervals. All minima are positive. (b) A stored witness certifies at $(\text{context }9,U37)$, while all three budget-limited rediscovery runs abstain. The 23 difficult cells were selected before execution.}
  \label{fig:reference-stability}
\end{figure}
Figure~\ref{fig:reference-stability} distinguishes certified non-uniqueness from finite-search incompleteness. For restart-seed order $i<j$, the archived statistic is $d_{\mathrm{rel}\ell_2}(a_i,a_j)=\lVert a_i-a_j\rVert_2/\max(\lVert a_i\rVert_2,10^{-12})$. Its asymmetric scale is descriptive; the claim rests on numerical distinction. Panel (a) shows cell-wise minima, medians, and maxima, ordered by context and Utility within each backbone. Each cell's 28 distances share eight actions and are dependent. Qwen has 16 cells with eight certified solutions each (448 pairs): global minimum $0.0225340$, maximum $1.1948683$, and minimum cosine similarity $0.2861448$. SmolLM2 adds eight cells, $64/64$ certified runs, and $224/224$ distinct pairs, with minimum distance $0.206739$. These diagnostics establish fixed-cell non-uniqueness on two backbones; they do not establish prevalence, connectivity, manifold structure, or global identifiability. Panel (b) compares stored-witness replay and budget-limited rediscovery on the same preselected cells. The $0/3$ result at $(\text{context }9,U37)$ is an existence counterexample, not a population failure-rate estimate. The distinct cell $(\text{context }9,U39)$ lacks a stored reference and is excluded.

\begin{figure}[H]
  \centering
  \includegraphics[width=\linewidth]{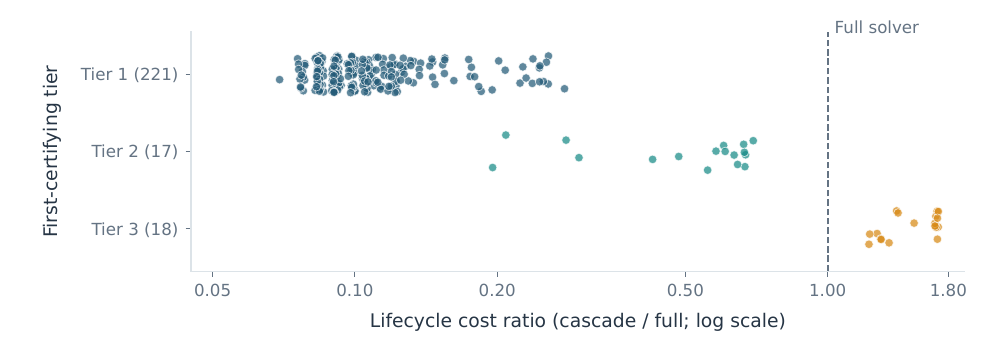}
  \caption{Singleton lifecycle cost by first-certifying tier on Qwen Frozen Fresh. Each dot is one request; vertical offsets separate overlapping dots, and parenthesized counts sum to 256. Costs include failed tiers. The dashed line marks full-solver cost: requests to its left are cheaper, while the Tier 3 tail extends beyond parity.}
  \label{fig:fresh-diagnostics}
\end{figure}

\begin{figure}[H]
  \centering
  \includegraphics[width=\linewidth]{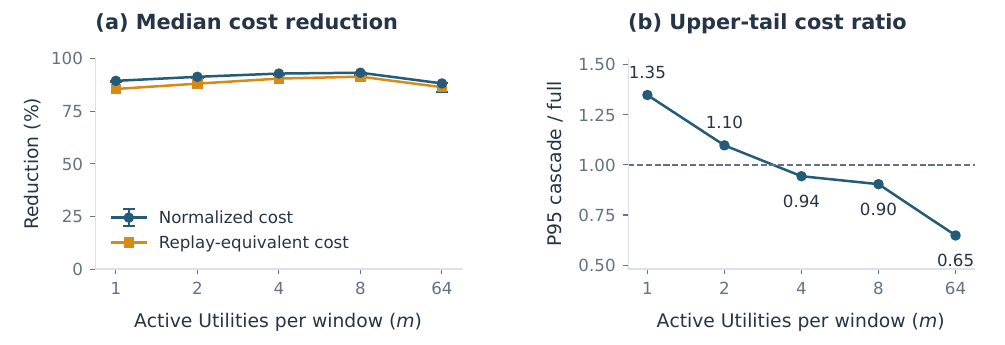}
  \caption{Window-level cost on Qwen Frozen Fresh. (a) Median cost reductions; downward whiskers show the protocol-defined 95\% bootstrap lower bounds for normalized cost. (b) P95 cascade/full ratio, with parity at one. Amortizing shared acquisition brings the tail below parity from $m=4$ onward; per-request search is unchanged. Window sizes are ordered conditions.}
  \label{fig:cost-frontier}
\end{figure}

\section{Extended Related Work and Positioning}
\label{app:related}
The closest proposal mechanisms differ mainly in what conditions them. HyperSteer conditions vector generation on language instructions and internal model state \citep{sun2025hypersteer}; DSEM conditions token-block vectors on episodic retrieval \citep{do2025dsem}; SVF constructs directions as local gradient fields \citep{li2026svf}; CLAS and IDS adjust strength \citep{hsu2026clas,vogels2026ids}; FineSteer jointly controls whether and how to steer \citep{weng2026finesteer}; per-instance multi-layer steering predicts where to intervene \citep{nelwan2026perinstance}; Steer2Adapt composes a reusable semantic basis for task adaptation \citep{han2026steer2adapt}; MoRe composes a latent-role codebook into a query-specific steering vector \citep{zeng2026more}; DiffuSteer learns a conditional generative proposal family in activation space \citep{wang2026diffusteer}; and ATLAS uses a learned hidden-state verifier to decide whether and how strongly to apply steering for each example and reasoning step \citep{nguyen2026atlas}. To our knowledge, these methods focus on proposal or selection rather than defining replay-certified membership and execution authorization as distinct runtime objects.

Non-identifiability and attribution work narrows what steering success alone can establish. Behaviorally indistinguishable vector classes undermine unique recovery of a steering object, while cross-encoding evaluation and SteerCheck show that efficacy need not establish encoding-invariant semantic attribution \citep{venkatesh2026nonidentifiability,gao2026crossencoding,luo2026steercheck}. The fixed-state experiment operationalizes a narrower analogue of this ambiguity under one frozen contract; ReplayCert serves admission rather than attribution.

The closest online-control methods differ in their assurance object. RE-Control optimizes a learned latent value \citep{kong2024recontrol}, PID Steering closes an activation-error loop \citep{nguyen2025pid}, and A-LQR solves feedback control over a local linearization \citep{skifstad2026alqr}. ObserverBench fixes the intervention contract and reports observer accuracy separately from chosen-action loss, linking mechanism estimates to the decisions they induce \citep{erramilli2026observerbench}. ReplayCert addresses a different boundary: a proposed neural action enters the current certified realization set only after isolated FP32/BF16 execution of the frozen model, and commitment additionally requires the run audit.
MERA calibrates error-reducing interventions and may abstain \citep{hedstrom2025mera}; BRT-Align uses learned reachability to anticipate unsafe latent trajectories \citep{karnik2025brtalign}; Neural Simplex and shielding provide system-level patterns for separating an advanced proposer from an assurance mechanism \citep{phan2020neuralSimplex,alshiekh2018shielding}.
Replay-gated execution does not replace the steering quality, guarantees, or safety objectives of these methods. Its narrower comparison axis is whether persistent behavioral specification, admissible realization set, search witness, and commitment authority are treated as distinct objects, and whether replay of known witnesses is evaluated separately from budget-limited operational rediscovery.

\end{document}